\documentclass[conference]{./support/ieeeconf}
\usepackage{times}
\usepackage{amsmath}
\usepackage[pdftex]{graphicx}
\usepackage{subfigure}
\usepackage{amsmath,amssymb,amsopn,amstext,amsfonts}
\usepackage{cancel}
\usepackage[space]{cite}
\usepackage{pdfsync}
\usepackage{balance}
\usepackage{color}
\usepackage{algorithm}
\usepackage{algorithmic}
\usepackage{url}
\usepackage{booktabs}
\usepackage{threeparttable}
\usepackage[linkcolor=black,citecolor=black,urlcolor=black,colorlinks=true]{hyperref}
\usepackage{multirow}
\usepackage[outdir=./epstopdf/]{epstopdf}
\usepackage{graphicx}
\IEEEoverridecommandlockouts

\newcommand\inv[1]{#1\raisebox{1.15ex}{$\scriptscriptstyle-\!1$}}
\DeclareMathOperator*{\argmaxA}{arg\,max} 
\DeclareMathOperator*{\argminA}{arg\,min}

\graphicspath{{./figure/}}
\DeclareGraphicsExtensions{.pdf,.png,.jpg,.eps}
\title{\LARGE \bf A General Optimization-based Framework for Global \\Pose Estimation with Multiple Sensors}
\author{Tong Qin, Shaozu Cao, Jie Pan, and Shaojie Shen
\thanks{All authors are with the Department of Electronic and Computer Engineering,
		Hong Kong University of Science and Technology, Hong Kong, China.
		{\tt\small \{tong.qin, shaozu.cao, jie.pan\}@connect.ust.hk, eeshaojie@ust.hk}.}
}

\begin{document}

\maketitle
\thispagestyle{empty}
\pagestyle{empty}

\begin{abstract}
Accurate state estimation is a fundamental problem for autonomous robots.
To achieve locally accurate and globally drift-free state estimation, multiple sensors with complementary properties are usually fused together.
Local sensors (camera, IMU, LiDAR, etc) provide precise pose within a small region, while global sensors (GPS, magnetometer, barometer, etc) supply noisy but globally drift-free localization in a large-scale environment.
In this paper, we propose a sensor fusion framework to fuse local states with global sensors, which achieves locally accurate and globally drift-free pose estimation. 
Local estimations, produced by existing VO/VIO approaches, are fused with global sensors in a pose graph optimization. 
Within the graph optimization, local estimations are aligned into a global coordinate.
Meanwhile, the accumulated drifts are eliminated. 
We evaluate the performance of our system on public datasets and with real-world experiments.
Results are compared against other state-of-the-art algorithms. 
We highlight that our system is a general framework, which can easily fuse various global sensors in a unified pose graph optimization.
Our implementations are open source\footnote{https://github.com/HKUST-Aerial-Robotics/VINS-Fusion}.

\end{abstract}

\section{Introduction}
Autonomous robot has become a popular research topic over the last decades.
We have seen an increasing demand for robots in various applications, such as autonomous driving, inspection, search and rescue.
One of the fundamental technologies for autonomous tasks is localization. 
Robots require precise 6-DoF (Degrees of Freedom) poses for navigation and control. 
A lot of sesnors have been used for local pose estimation.
The Radar and LiDAR are widely used in the confined indoor environment, while the camera and IMU are applicable in both indoor and outdoor environments.
There are many impressive algorithms for local pose estimation, such as visual-based method \cite{klein2007parallel, ForPizSca1405, engel2014lsd, mur2015orb, engel2017direct}, and visual-inertial-based method \cite{MouRou0704,LiMou1305,LeuFurRab1306,bloesch2015robust,mur2017visual,qin2018vins}. 
These algorithms achieves incremental and accurate state estimation within a local region.
However, there are several drawbacks limiting the usage of these algorithms in practice.

The first drawback of local pose estimation algorithems is that they produce pose estimation in a local frame (with respect to the starting point) without a global coordinate. 
We may get different estimations when we start from different points even in the same environment. 
Hence, they are unfriendly to reuse without a fixed global coordinate. 
The second drawback is that due to the lack of global measurements, the local estimations are prone to accumulated drifts in the long run. 
Although some vision-based loop closure methods were proposed to eliminate drifts, they cannot handle the large-scale environment with the mass data.

\begin{figure}
	\centering
	\includegraphics[width=0.45\textwidth]{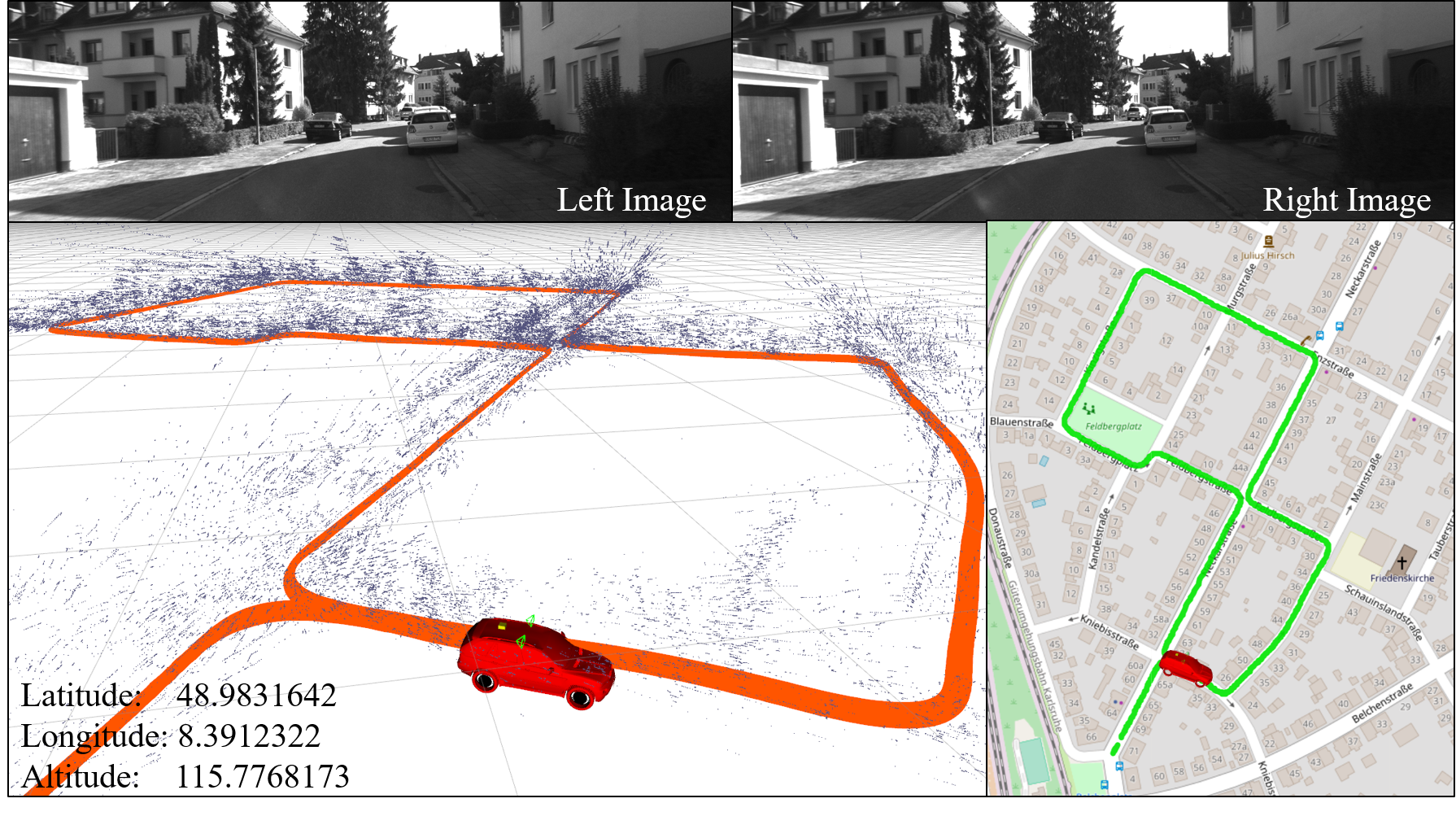}
	\caption{
		\label{fig:a_graph} KITTI dataset results of the proposed sensor fusion framework (VO + GPS). The top of this figure is a pair of stereo images. The left bottom part is the estimated trajectory and feature points, while the right bottom part is the estimated global trajectory aligned with Google map.}
\end{figure}

\begin{figure*}
	\centering
	\includegraphics[width=0.98\textwidth]{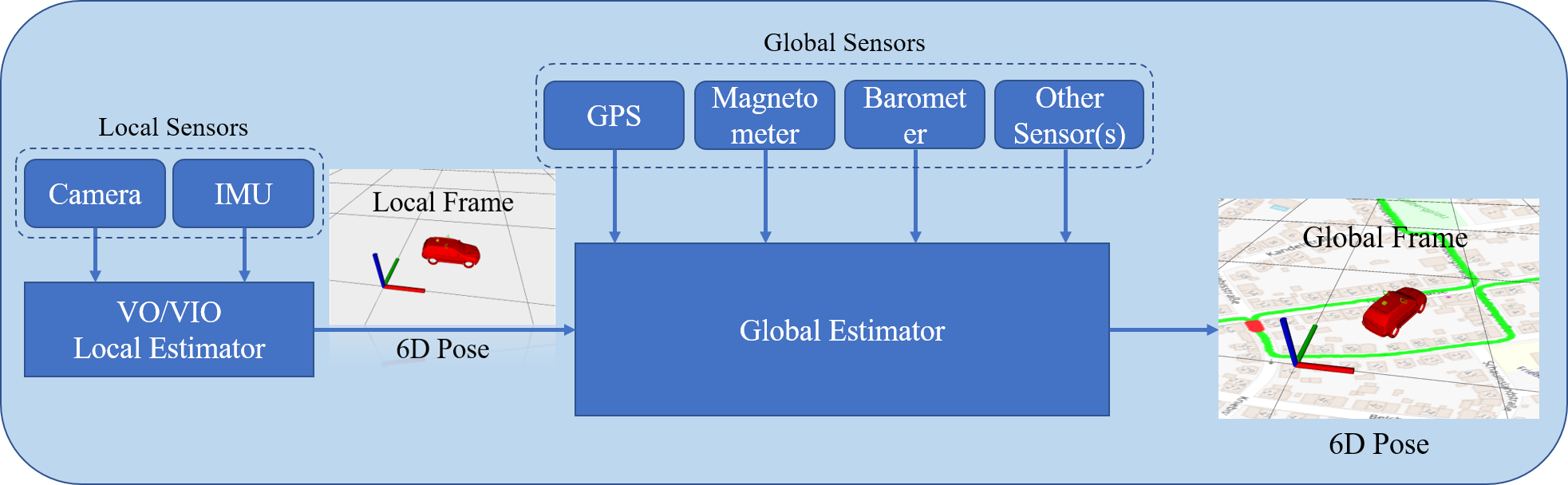}
	\caption{
		\label{fig:paperb_framework} 
		An illustration of the proposed framework structure. The global estimator fuses local estimations with various global sensors to achieve locally accurate and globally drift-free pose estimation. }
\end{figure*}

Compared with local sensors, global sensors, such as GPS, barometers, and magnetometers, have advantages in global localization within large-scale environments.
They provide global measurements with respect to the fixed earth frame, which is drift-free.
However, their measurements are usually unsmooth and noisy, which cannot be directly used for precise control and navigation. 
Taking GPS as an example, it can measure approximate location in meters, but  measurements are discontinuous at a low rate.
Also, it only measures 3D position without 3D orientation. 
Therefore, only global sensors are insufficient for real-time 6-DoF state estimation.

Since local sensors (camera, IMU and LiDAR) achieves impressive performance in local accuracy and global sensors (GPS, magnetometer and barometer) are drift-free, it is a smart way to fuse them together to achieve locally accurate and globally drift-free 6-DoF pose estimation. 
In order to increase the robustness, we want to fuse sensors as many as possible. 
Consequently, a general framework which supports multiple sensors is required.
Although traditional EKF-based methods can fuse the local estimation into the global frame gradually, an accurate initial guess about the transformation between different frames is required to guarantee convergence. 
Also, the EKF methods are sensitive to time synchronization. 
Any late-coming measurements will cause trouble since states cannot be propagated back in filter procedure.
To this end, we use an optimization-based method to solve this problem, which is suitable for multiple sensor fusion inherently.

In this paper, we propose an optimization framework to fuse local estimations with global sensor measurements. 
Local estimations come from existing state-of-the-art VO/VIO works.
Global sensors are treated as general factors in pose graph.
Local factors and global sensor factors are summed up together to build the optimization problem.
Our system achieves locally accurate and globally drift-free state estimation. 
We highlight the contribution of this paper as follows:
\begin{itemize}
	\item a general framework to fuse various global sensors with local estimations, which achieves locally accurate and globally drift-free localization.
	\item an evaluation of the proposed system on both public datasets and real experiments. 
	\item open-source code for the community.
\end{itemize}

\section{Related Work}
\label{sec:literature}
Recently, multiple sensor fusion approach for state estimation has become a popular trend in order to improve both accuracy and robustness. 
By the type of sensors employed in the system, research works can be classified as local localization and global-aware localization.

For local localization, cameras, IMU, LiDAR and RGB-D sensors are usually used for 6-DoF state estimation in small environment.
Impressive approaches over the last decades include visual-based methods \cite{klein2007parallel, ForPizSca1405, engel2014lsd, mur2015orb, engel2017direct}, LiDAR-based methods \cite{zhang2014loam}, RGB-D based methods \cite{kerl2013dense}, and event-based methods \cite{rebecq2017evo}. 
There are also some multi-sensor fusion methods, such as visual-inertial fusion \cite{MouRou0704,LiMou1305,huang2010observability,LeuFurRab1306,bloesch2015robust,mur2017visual,qin2018vins,liu2018ice} and visual-LiDAR fusion \cite{zhang2015visual}.
Among these work, \cite{MouRou0704,LiMou1305,bloesch2015robust} are filter-based methods while \cite{LeuFurRab1306,mur2017visual,qin2018vins} are optimization-based methods. 
In optimization-based framework, a lot of visual measurements and inertial measurements are kept in a bundle.
The states related to the observed measurements are optimized together.
One advantage of optimization-based approaches over EKF-based ones is that states can be iteratively linearized to increase accuracy.
Both filter-based methods and optimization-based methods can achieve highly accurate state estimation. 
Due to the lack of global measurement, the accumulated drifts is unavoidable over time.

For global-aware localization, global sensors (GPS, magnetometer, barometer, etc) are incorporated in the system.
Global sensor measures absolute quantities with respect to the earth frame, which are independent of the starting point.
Global measurements are usually noisy and low-frequency, so they cannot be used alone. 
Therefore, global sensors are usually fused with local sensors for accurate and global-aware localization.
\cite{lynen2013robust} proposed an EKF-based algorithm to fuse visual measurement with inertial and GPS measurement to get drift-free estimation.
\cite{SheMulMic1405} used UKF (Unscented Kalman Filter) algorithm to fuse visual, LiDAR and GPS measurements, which is an extension of EKF without analytic Jacobians. 
Filter-based methods are sensitive to time synchronization. 
Any late-coming measurements will cause trouble since states cannot be propagated back in filter procedure. 
Hence, special ordering mechanism is required to make sure that all measurements from multiple sensors are in order. 
Compared with filter-based method, optimization-based method have advantage in this aspect. 
Because the big bundle serves as a nature buffer, it can wait and store measurements for a long time.
\cite{mascaro2018gomsf} used an optimization-based framework to fuse local VIO (Visual Inertial Odometry) with GPS measurements, which produced more accurate results than method proposed in \cite{lynen2013robust}.
The transformation between local coordinate and global coordinate was frequently optimized in this approach. 
Few research works fuse sensors more than three types.
In this paper, we propose a more general optimization-based framework for global localization, which can support multiple global sensors.
Each sensor serves as a general factor, which can be easily added into the optimization problem.

\section{System Overview}
\label{sec:System Overview}

According to the measurement's reference frame, we category sensors into local and global types.
\subsubsection{Local Sensors}
Camera, LiDAR, IMU (accelerometer and gyroscope), etc.
This kind of sensor is not globally referenced, thus a reference frame is usually needed. 
In general, the first pose of the robot is set as the origin in order to boot up the sensor.
The estimation of the robot's pose incrementally evolves from the starting point.
Therefore, accumulated drift will grow with the distance from starting point.  

\subsubsection{Global Sensors}
GPS, magnetometer, barometer, etc.
This kind of sensor is globally referenced.
It always works under a fixed global frame, such as the earth frame.
The origin of the reference frame is fixed and known in advance. 
Their measurements are global-referenced but noisy.
The error is independent of traveled distance. 
For GPS, it measures absolute longitude, latitude and altitude with respect to the earth. 
The longitude, latitude, and altitude can be converted to x, y and z coordinate.
For magnetometer, it measures magnetic field direction and strength, which can determine the orientation.
For barometer, it measures air pressure, which can be converted to height.

The structure of our framework is shown in Fig. \ref{fig:paperb_framework}.
Local sensors (camera and IMU) are used in local estimation. 
The existing VO/VIO approaches are adopted to produce local poses. 
Local results and global sensors are input into a global pose graph. 
They are converted to unified factors to construct the optimization problem.  
The global estimator generates locally accurate and globally aware 6-DoF pose results.

\section{Methodology}
\label{sec:algorithm}
\subsection{Local Pose Estimation}
For local pose estimation, we adopt existing VO (Visual Odometry)/VIO (Visual-Inertial Odometry) algorithms.
There are many impressive VO/VIO algorithms, such as \cite{mur2015orb, MouRou0704,LiMou1305,LeuFurRab1306,bloesch2015robust,mur2017visual,qin2018vins}. 
Any of them can be used as local pose estimation in our framework, as long as it produces 6-DoF poses. 
This part is not the main contribution of this paper.
For the completeness, we briefly introduce our previous VIO algorithm \cite{qin2018vins}, which is used in our open-source implementations. 

The VIO estimates poses of several IMU frames and features' depth within a sliding window.
The states are defined as:
\begin{equation}
\begin{split}
\mathcal{X}_l    &= \left [ \mathbf{x}_0,\,\mathbf{x}_{1},\, \cdots \,\mathbf{x}_{n},\,  \lambda_0,\,\lambda_{1},\, \cdots \,\lambda_{m} \right ] \\
\mathbf{x}_k   &= \left [ \mathbf{p}^l_{b_k},\,\mathbf{v}^l_{b_k},\,\mathbf{q}^l_{b_k}, \,\mathbf{b}_a, \,\mathbf{b}_g \right ], k\in [0,n],
\end{split}
\end{equation}
where the $k$-th IMU state $\mathbf{x}_k$ consists of the position $\mathbf{p}^{l}_{b_k}$, velocity $\mathbf{v}^{l}_{b_k}$, orientation $\mathbf{q}^{l}_{b_k}$ of IMU's center with respect to local reference frame $l$. 
We use quaternion to represent orientation.
The first IMU pose is set as reference frame.
$\mathbf{b}_a$ and $\mathbf{b}_g$ are accelerometer bias and gyroscope bias respectively. 
Features are parameterized by their inverse depth $\lambda$ when first observed in camera frame.
The estimation is formulated as a nonlinear least-squares problem,

\begin{equation}
\label{eq:nonlinear_cost_function}
\begin{aligned}
\min_{\mathcal{X}_l} \left \{ \left 
\| \mathbf{r}_p - \mathbf{H}_p \mathcal{X} \right \|^2 
+ \sum_{k \in \mathcal{B}}  \left \| \mathbf{r}_{\mathcal{B}}(\hat{\mathbf{z}}^{b_k}_{b_{k+1}},\, \mathcal{X}) \right \|_{\mathbf{P}^{b_k}_{b_{k+1}}}^2 + \right. \\
\left. 
\sum_{(l,j) \in \mathcal{C}} \rho( \left \| \mathbf{r}_{\mathcal{C}}(\hat{\mathbf{z}}^{c_j}_l ,\, \mathcal{X}) \right \|_{\mathbf{P}^{c_j}_l}^2 )
\right\},
\end{aligned}                    
\end{equation}
where $\mathbf{r}_{\mathcal{B}}(\hat{\mathbf{z}}^{b_k}_{b{k+1}},\, \mathcal{X})$ and $\mathbf{r}_{\mathcal{C}}(\hat{\mathbf{z}}^{c_j}_l,\, \mathcal{X})$ represent inertial and visual residuals respectively. 
The prior term, $\{\mathbf{r}_p,\,\mathbf{H}_p\}$, contains information about past marginalized states.
$\rho (\cdot)$ represents robust huber norm \cite{Hub64}.
The detailed explanation can be found at \cite{qin2018vins}. 
The VIO achieves accurate real-time 6-DoF pose estimations in the local frame.

\begin{figure}
	\centering
	\includegraphics[width=0.4\textwidth]{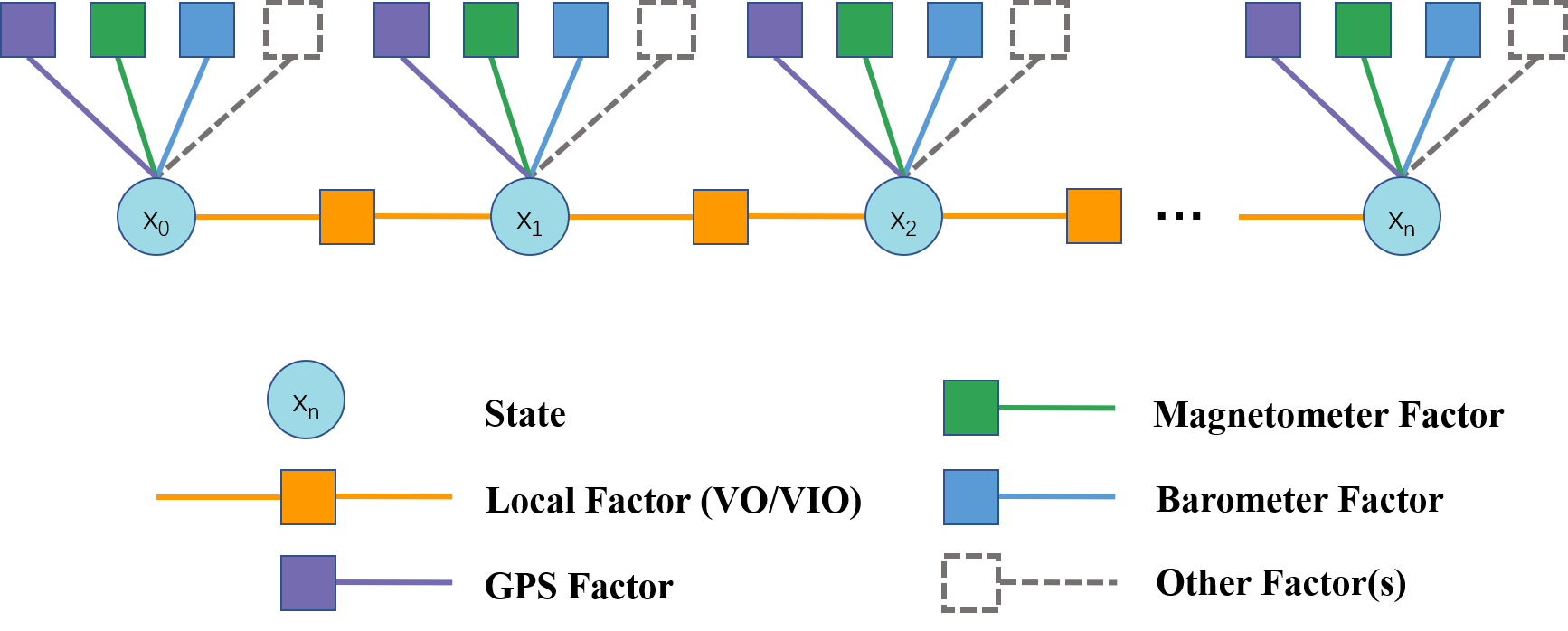}
	\caption{
		\label{fig:factor_graph} 
		An illustration of the global pose graph structure. 
		Every node represents one pose in world frame, which contains position and orientation.
		The edge between two consecutive nodes is a local constraint, which is from local estimation (VO/VIO). 
		Other edges are global constraints, which come from global sensors.
	}
\end{figure}

\subsection{Global Pose Graph Structure}

The illustration of the global pose graph structure is shown in Fig. \ref{fig:factor_graph}.
Every pose, which contains position and orientation in world frame, serves as one node in the pose graph.
The density of nodes is determined by the lowest-frequency sensor.
The edge between two consecutive nodes is a local constraint, which is from local estimation (VO/VIO). 
That edge constrains the relative pose from one node to another node.
Other edges are global constraints, which come from global sensors.

The nature of pose graph optimization is an MLE (Maximum Likelihood Estimation) problem. 
The MLE consists of the joint probability distribution of robot poses over a time period.
Variables are global poses of all nodes, $\mathcal{X}=\{\mathbf{x}_0, \mathbf{x}_1, ..., \mathbf{x}_n\}$, where $\mathbf{x}_i = \{\mathbf{p}^w_i, \mathbf{q}^w_i\}$.
$\mathbf{p}^w$ and $\mathbf{q}^w$ are position and orientation under the global frame. 
Under the assumption that all measurement probabilities are independent, the problem is typically derived as,
\begin{equation}
\mathcal{X}^* = \argmaxA_{\mathcal{X}}  \prod^{n}_{t=0}\prod^{}_{k\in{\mathbf{S}}} p(\mathbf{z}^k_t|\mathcal{X}),
\end{equation}
where $\mathbf{S}$ is the set of measurements, which includes local measurements (VO$\backslash$VIO) and global measurments (GPS, magnetometer, barometer and so on).
We assume the uncertainty of measurements are Gaussian distribution with mean and covariance, which is  $p(\mathbf{z}^k_t|\mathcal{X}) \sim \mathcal{N}(\tilde{\mathbf{z}}^k_t, \Omega^k_t)$.
Therefore, the above-mentioned equation is derived as, 
\begin{equation}
\label{eq:ba}
\begin{split}
\mathcal{X}^* &= \argmaxA_{\mathcal{X}}  \prod^{n}_{t=0} \prod^{}_{k\in{\mathbf{S}}} exp(-\frac{1}{2}\left\|\mathbf{z}^k_t - h^k_t(\mathcal{X})\right\|^2_{ \mathbf{\Omega}^k_t}) \\
&= \argminA_{\mathcal{X}} \sum^n_{t=0} \sum^{}_{k\in{\mathbf{S}}}
\left\|\mathbf{z}^k_t - h^k_t(\mathcal{X})\right\|^2_{ \mathbf{\Omega}^k_t}.
\end{split}
\end{equation}
The Mahalanobis norm is $\left \| \mathbf{r} \right\|_\mathbf{\Omega}^2 = \mathbf{r}^T  \mathbf{\Omega}^{-1}{\mathbf{r}}$. 
Then the state estimation is converted to a nonlinear least squares problem, which is also known as Bundle Adjustment (BA).

\subsection{Sensor Factors}

\subsubsection{Local Factor}
Since the local estimation (VO/VIO) is accurate within a small region, we take advantage of the relative pose between two frames.
Considering two sequential frame $t-1$ and frame $t$, the local factor is derived as,

\begin{equation}
\begin{split}
\mathbf{z}^{l}_t - h^{l}_t(\mathcal{X}) &= \mathbf{z}^{l}_t - h^{l}_t(\mathbf{x}_{t-1},\mathbf{x}_{t}) \\
&=
\begin{bmatrix}
\inv{\mathbf{q}^{l}_{t-1}}(\mathbf{p}^l_t - \mathbf{p}^l_{t-1})  \\
\inv{\mathbf{q}^{l}_{t-1}} {\mathbf{q}^{l}_{t}}
\end{bmatrix}
\ominus
\begin{bmatrix}
\inv{\mathbf{q}^{w}_{t-1}}(\mathbf{p}^w_t - \mathbf{p}^w_{t-1})  \\
\inv{\mathbf{q}^{w}_{t-1}} {\mathbf{q}^{w}_{t}}
\end{bmatrix},
\end{split}
\end{equation}
where ($\mathbf{q}^l_{t-1}, \mathbf{p}^l_{t-1})$ and ($\mathbf{q}^l_{t}, \mathbf{p}^l_{t})$ are poses at time $t-1$ and $t$ in local frame from VO/VIO. 
$\ominus$ is the minus operation on the error state of quaternion.
The first row represents relative position error between two poses, and the second row represents relative rotation error between two poses. 
If the VO/VIO algorithm produces the covariance matrix of poses, we use it as the covariance of local measurements.
Otherwise, we use the unified covariance for all local measurments.

\subsubsection{GPS Factor}
Raw measurements of GPS are longitude, latitude, and altitude, which are not in x,y, and z-axis coordinates. 
Generally, we can convert longitude, latitude and altitude into ECEF (Earth Centred Earth Fixed), ENU (local East North Up), and NED (local North East Down) coordinates.
Here, we take ENU coordinate as the example.
By setting the first GPS measurement as the origin point, we get GPS's measruments in the ENU world frame, $\mathbf{p}^{GPS}_t = [x^w_t, y^w_t, z^w_t]^T$.
The GPS factor is derived as,
\begin{equation}
\mathbf{z}^{GPS}_t - h^{GPS}_t(\mathcal{X}) 
= \mathbf{z}^{GPS}_t - h^{GPS}_t({\mathbf{x}_t})
= \mathbf{p}^{GPS}_t - \mathbf{p}^w_t .
\end{equation}
The GPS measurements directly constrain the position of every node. 
The covariance is determined by the number of satellites when the measurement is received. 
The more satellites it receives, the smaller covariance it is.

\subsubsection{Magnetometer Factor}
The magnetometer can measure a vector of the magnetic field intensity. 
The direction of this vector can help to determine orientation in the world frame. 
We assume the magnetometer is calibrated offline without offset or bias.
First of all, we lookup table to get the magnetic intensity $\mathbf{z}^w$ of the local region in the ENU coordinate.
We assume that the magnetic intensity $\mathbf{z}^w$ is constant within this area. 
Our measurement is denoted as $\mathbf{z}^{m}_t$.
The orientation of $\mathbf{z}^{m}_t$ should match $\mathbf{z}^w$ if we put the sensor coinciding with the ENU coordinate.
Inspired by this, the factor is derived as:
\begin{equation}
\begin{split}
\mathbf{z}^{m}_t - h^{m}_t(\mathcal{X}) 
&=\mathbf{z}^{m}_t - h^{m}_t(\mathbf{x}_t) \\
&=\frac{\mathbf{z}^{m}_t }{\|\mathbf{z}^{m}_t\|} - 
\mathbf{q}^m_b \inv{\mathbf{q}^w_t}\frac{ \mathbf{z}^w }{\| \mathbf{z}^w\|},
\end{split}
\end{equation}
where $\mathbf{q}^m_b$ is the transformation from robot's center to the magnetometer's center, which is known and calibrated offline.
Since the magnetic field is easily affected by the environment, we only use the normalized vector without length.
The length is used to determine covariance. 
If the length of measurement differs $\mathbf{z}^w$ a lot, we set a big covariance.
Otherwise, we use a small covariance. 

\subsubsection{Barometer Factor}

The barometer measures the air pressure in an area.
We assume that the air pressure is constant at one altitude over a period of time.
So the air pressure can be converted to height linearly.
As the same as GPS, we set the first measurement as the origin height.
Then we get the measurement of height $z^m_t$.
Intuitively, the factor is a residual of height estimation, which is written as:
\begin{equation}
\mathbf{z}^{m}_t - h^{m}_t(\mathcal{X}) = \mathbf{z}^{m}_t - h^{m}_t(\mathbf{x}_t) =
{z}^{m}_t - z_t .
\end{equation}
Since this measurement is noisy, we caculate the variance of several measurments within a short time, and use it in the cost function.

\subsubsection{Other Global Factors}
Though we only specify GPS, magnetometer, and barometer factors in detail, our system is not limited to these global sensors.
Other global sensors and even some artificial sensors, such as Motion Capture system, WiFi and Bluetooth fingerprint, can be used in our system. 
The key is to model these measurements as residual factors under one global frame. 

\subsection{Pose Graph Optimization}
Once the graph is built, optimizing it equals to finding the configuration of nodes that match all edges as much as possible. 
Ceres Solver~\cite{ceres-solver} is used for solving this nonlinear problem, which utilizes Gaussian-Newton and Levenberg-Marquadt approaches in an iterative way. 

We run pose graph optimization at low frequency (1Hz). 
After every optimization, we get the transformation for local frame to global frame. 
Therefore, We can transform subsequent high-rate local poses (VO/VIO, 200Hz) by this transformation to achieve real-time high-rate global poses.
Since the pose graph is quite sparse, the computation complexity increases linearly with the number of poses.
We can keep a huge window for pose graph optimization to get accurate and globally drift-free pose estimation. 
When the computation complexity exceeds real-time capability, we throw old poses and measurements, and keep the window in a limited size.

\section{Experimental Results}
\label{sec:experiments}
We evaluate the proposed system with visual and inertial sensors both on datasets and with real-world experiments.  
In the first experiment, we compare the proposed algorithm with another state-of-the-art algorithm on public datasets. 
We then test our system with self-developed sensor suite in the real-world outdoor environment.
The numerical analysis is generated to show the accuracy of our system in detail.

\subsection{Datasets}

We evaluate our proposed system using KITTI Datasets~\cite{geiger2012we}. 
The datasets are collected onboard a vehicle, 
which contains stereo images (Point Grey Flea 2, 1382x512 monochrome, 10 FPS) and GPS. 
Images are synchronized and rectified.
The transformation between sensors is calibrated.
Also, the ground truth states are provided by the Inertial Navigation System (OXTS RT 3003).
We run datasets with stereo cameras, latitude, longitude, and altitude from raw GPS measurements.
The stereo cameras are used for local state estimation. 
The local results are fused with GPS measurements in global optimization.

\begin{figure}
	\centering
	\includegraphics[width=0.45\textwidth]{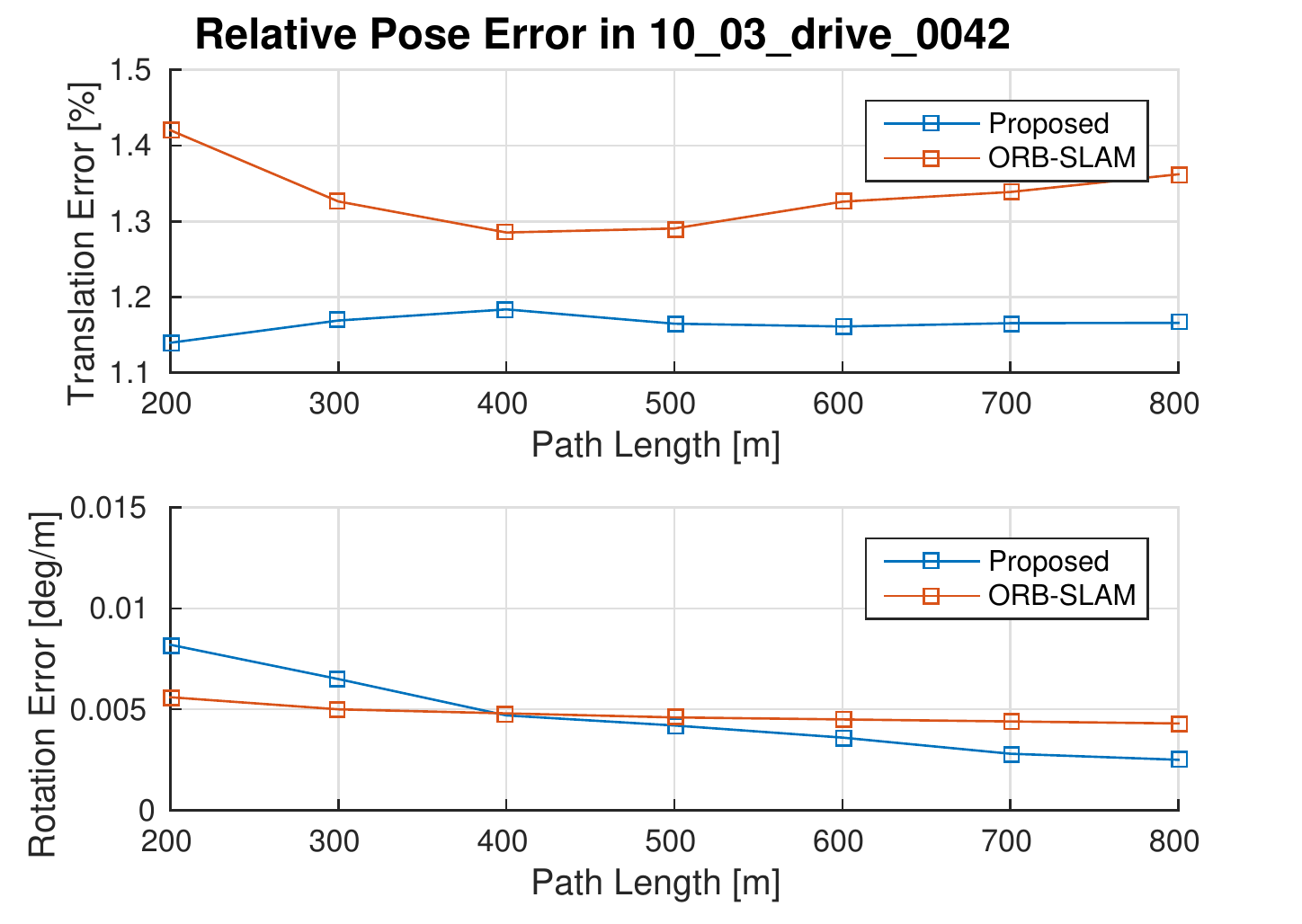}
	\caption{
		\label{fig:kitti01} Rotation error and translation error plot in 10\_03\_drive\_0042.}
\end{figure}

\begin{figure}
	\centering
	\includegraphics[width=0.45\textwidth]{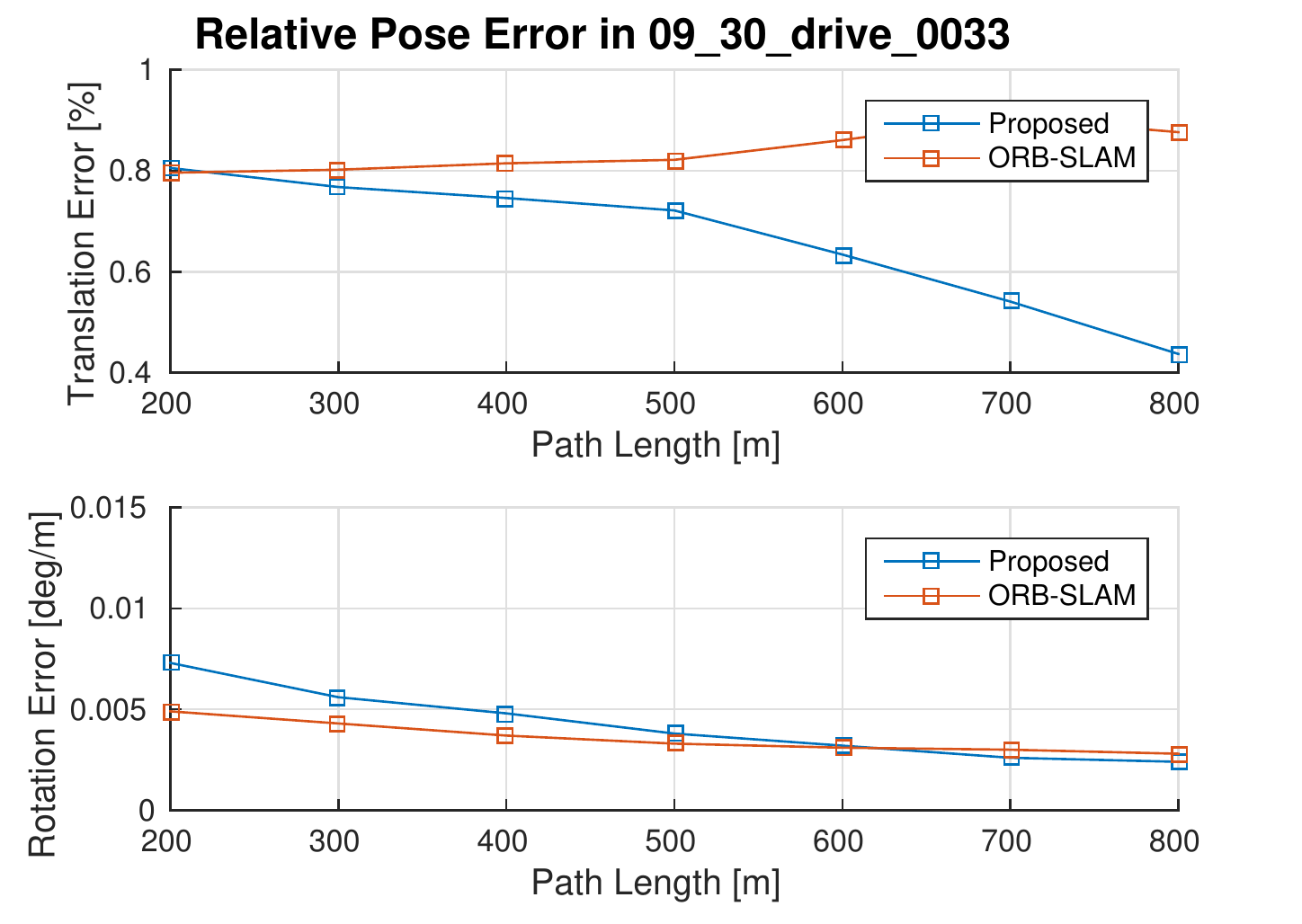}
	\caption{
		\label{fig:kitti09} Rotation error and translation error plot in 09\_30\_drive\_0033.}
\end{figure}

\begin{figure}
	\centering
	\includegraphics[width=0.4\textwidth]{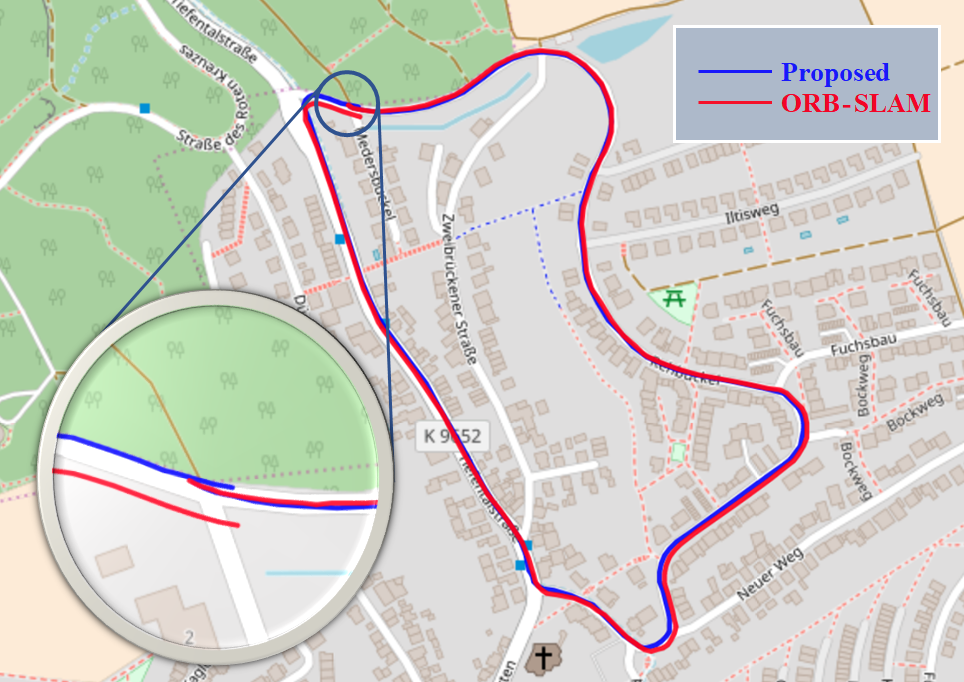}
	\caption{
		\label{fig:propose_ORB} Trajectories of one KITTI sequence (09\_30\_drive\_0033) recovered from ORB-SLAM and the proposed algorithm. 
	}
\end{figure}

\begin{table}
	\centering
	\caption{ RMSE\cite{sturm2012benchmark} in KITTI datasets in meters.\label{tab:kitti_error}}
	\begin{tabular}{cccc}
		\toprule
		\multirow{2}{*}{Sequences}  &
		\multirow{2}{*}{Length[km] } &
		\multicolumn{2}{c}{RMSE[m]} 
		 \\
		\cline{3-4}
		& & ORB-SLAM &  Proposed \\
		\midrule
		09\_30\_drive\_0016      & 0.39  & 0.18 & \textbf{0.12}  \\
		09\_30\_drive\_0018     & 2.21  & 0.83 & \textbf{0.24}  \\
		09\_30\_drive\_0020       & 1.23  & 0.71 & \textbf{0.27}  \\
		09\_30\_drive\_0027     & 0.69  & 0.57 &  \textbf{0.15}  \\
		09\_30\_drive\_0033         & 1.71  & 3.01 &  \textbf{0.27}  \\
		09\_30\_drive\_0034        & 0.92  & 1.04 & \textbf{0.20}  \\
		10\_03\_drive\_0027         & 3.72  & 1.25 & \textbf{0.28}  \\
		10\_03\_drive\_0042       & 2.45  & 12.48 & \textbf{0.63}  \\
		\bottomrule
	\end{tabular}
\end{table}

In this experiment, we compare our results with ORB-SLAM~\cite{mur2015orb}, a state-of-the-art visual odometry method that works with stereo cameras. 
Orb-slam is an optimization-based algorithm with powerful relocalization capability. 
It maintains a map of keyframes and landmarks.
We evaluate RPE (Relative Pose Errors) and ATE (Relative Trajecotry Error) one results produced by proposed method and ORB-SLAM.  
The RPE is caculated by the tool proposed in \cite{geiger2012we}.
The position and rotation RPE of two sequences are shown in Fig. \ref{fig:kitti01} and Fig. \ref{fig:kitti09} respectively.
For translation error, the proposed method is obviously lower than ORB-SLAM. 
It shows that the position drift is effectively eliminated by GPS measurement. 
For rotation error, the proposed method is not better than ORB-SLAM.
Because GPS can't directly measure rotation angles, it cannot improve local accuracy on rotation.

The RMSE (Root Mean Square Errors) of absolute trajectory error for more sequences in KITTI datasets is shown in Table.~\ref{tab:kitti_error}. 
Estimated trajectories are aligned with the ground truth by Horn's method \cite{horn1987closed}.
For all sequences, the proposed method outperforms orb-slam, which demonstrates that fusing GPS measurements 
effectively increases the accuracy of the estimated trajectory.
Intuitively, GPS corrects accumulated drifts in the long run. 

Trajectories of one KITTI sequence (09\_30\_drive\_0033) recovered from ORB-SLAM and the proposed algorithm are shown in Fig. \ref{fig:propose_ORB}. 
Trajectories are aligned with Google map from the bird-eye view. 
From the picture, we can see that the estimated trajectory of ORB-SLAM drifts several meters at the end.
The estimated trajecotry of proposed method matches the road network well.
This experiment demonstrates that proposed system has advantage of pose estimation in a long distance.

\subsection{Real-world experiment}

\begin{figure}
	\centering
	\includegraphics[width=0.4\textwidth]{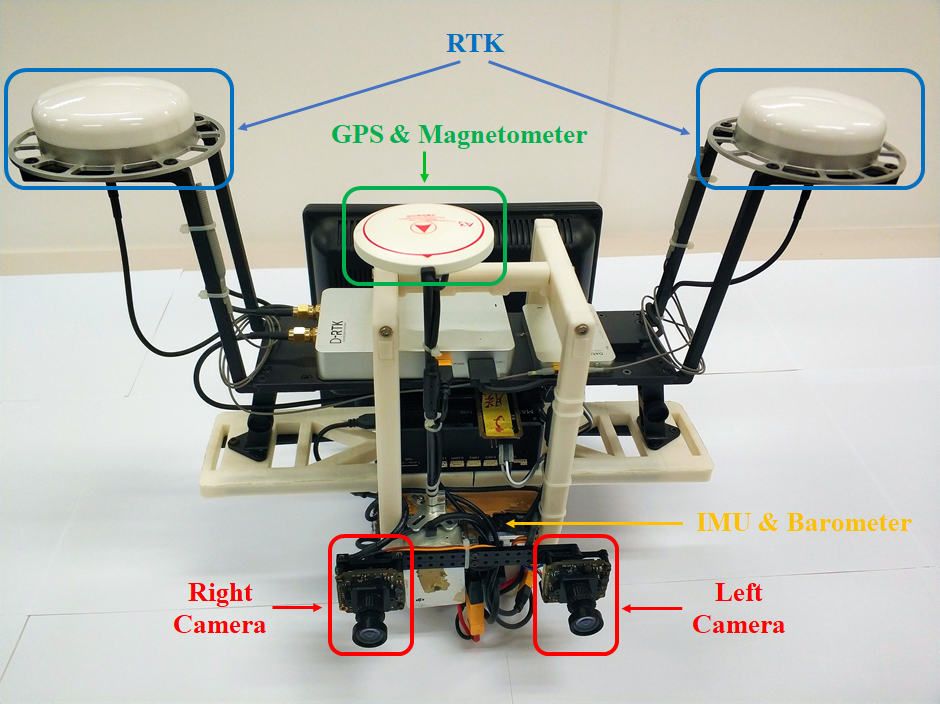}
	\caption{\label{fig:device} The sensor suite we used for the outdoor experiment, which contains two forward-looking global shutter cameras (MatrixVision mvBlueFOX-MLC200w) with 752x480 resolution. We use the built-in IMU, magnetometer, barometer and GPS from the DJI A3 flight controller.}
\end{figure}

\begin{figure}
	\centering
	\includegraphics[width=0.45\textwidth]{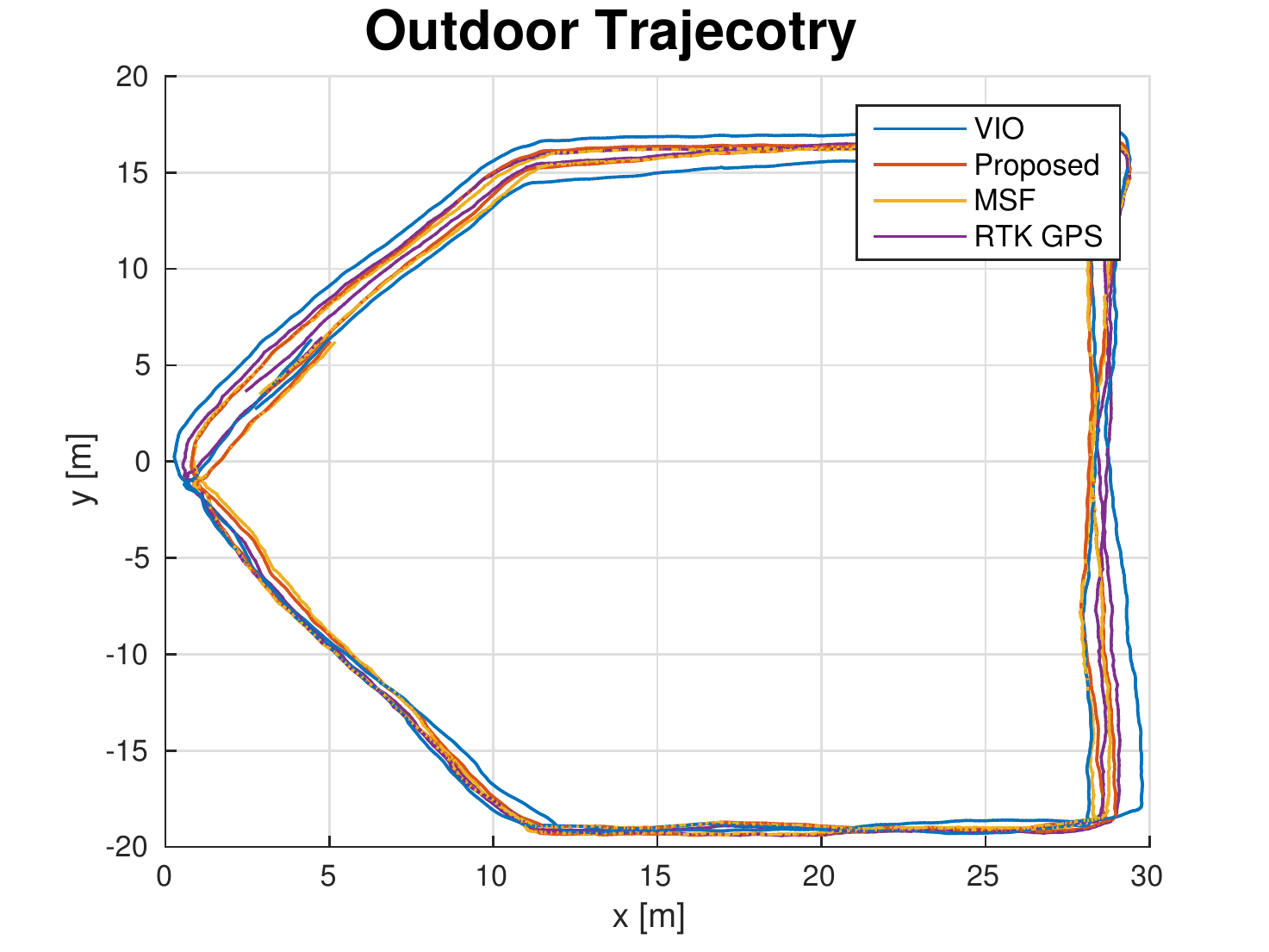}
	\caption{
		\label{fig:trajectory_b} The trajectories of our small-scale outdoor experiment recovered from VIO, the proposed algorithm and MSF respectively. We use RTK trajectory as ground truth.}
\end{figure}

\begin{figure}
	\centering
	\includegraphics[width=0.45\textwidth]{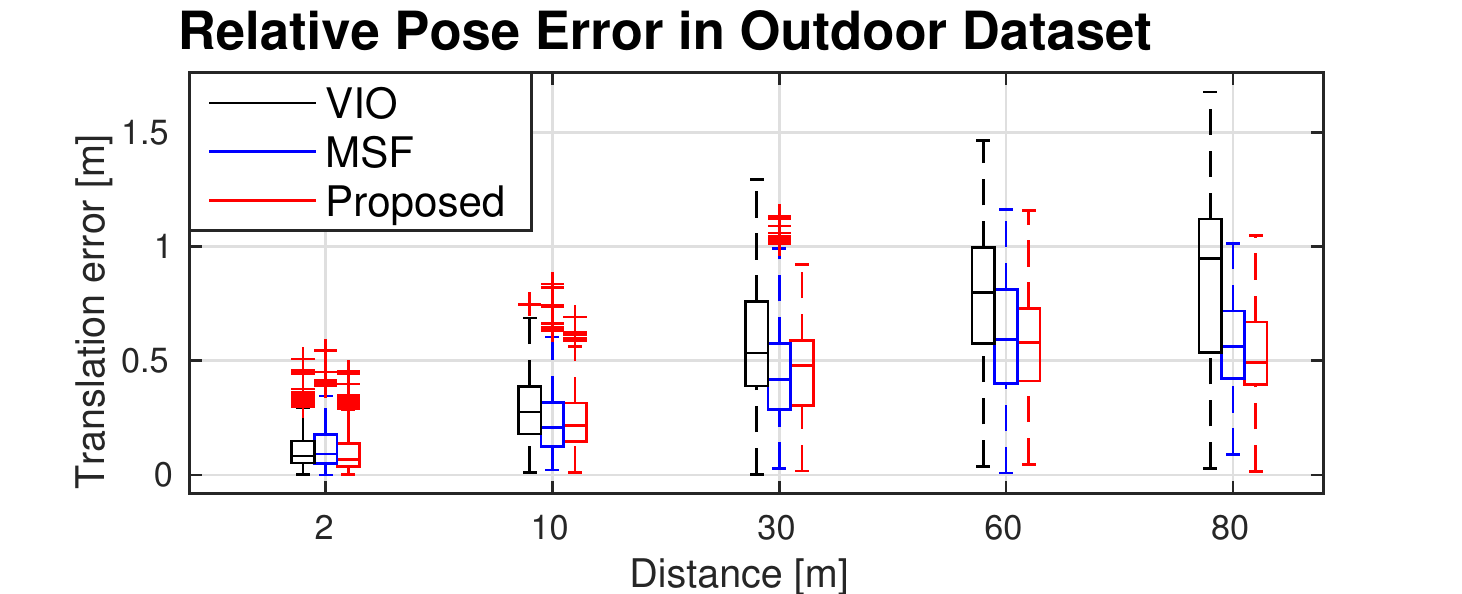}
	\caption{
		\label{fig:rpe_b} Relative pose error plot in our small scale outdoor experiment.}
\end{figure}

\begin{table}
	\centering
	\caption{{RMSE[m] in outdoor experiment.} \label{tab:outdoor_experiment}}
	\begin{tabular}{c c c  c c}
		\toprule
		\multirow{2}{*}{Sequence} & \multirow{2}{*}{Length[m]}  &
		\multicolumn{3}{c}{RMSE[m]}  \\
		\cline{3-5}
		&     &   VIO & Proposed & MSF \\
		\hline
		Outdoor1 & 242.05 & 0.77 & \textbf{0.40} & 0.66 \\
		Outdoor2 & 233.89 & 0.66 & \textbf{0.41} & 0.58 \\
		Outdoor3 & 232.13 & 0.75 & \textbf{0.38} & 0.63  \\
		\bottomrule
	\end{tabular}
\end{table}

\begin{figure}
	\centering
	\includegraphics[width=0.4\textwidth]{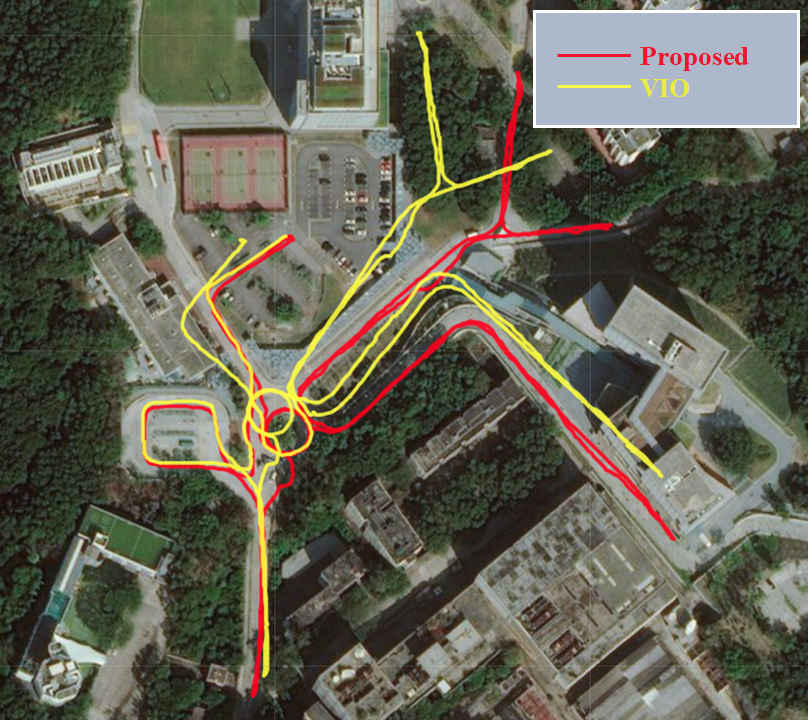}
	\caption{
		\label{fig:pathonmap} The trajectories of our large-scale outdoor experiment recovered from VIO and the proposed algorithm (VIO + GPS + magnetometer + barometer) respectively.}
\end{figure}

In this experiment, we used self-developed sensor suite which were equipped with multiple sensors.
The sensor suite is shown in Fig.~\ref{fig:device}.
It contains stereo cameras (mvBlueFOX-MLC200w, 20Hz) and DJI A3 controller\footnote{\url{http://www.dji.com/a3}}, which inculdes built-in IMU, magnetometer, barometer and GPS receiver. 
We also equip it with RTK (Real-Time Kinematic)\footnote{\url{https://www.dji.com/d-rtk}} receiver for high-accurate localization.
The RTK base station is established on the top of a building. 
The RTK can provide precise positioning of one centimeter vertically and horizontally, which is treated as ground truth.
We run states estimation with all available sensors.

For accuracy comparison, we walked two circles on the ground. 
We compare our results against MSF~\cite{lynen2013robust}, which fuses visual odometry, inertial measurements, and GPS in an EKF-based framework. 
The trajecotry comparison is shown in Fig.~\ref{fig:trajectory_b}, and the RPE (Relative Pose Error) is plotted in Fig.~\ref{fig:rpe_b}. 
We can see obvious translation drift in the estimated trajectory of VIO.
From the relative pose error, we can see that the proposed system improved the accuracy of VIO a lot.
Also, the proposed system outperforms MSF~\cite{lynen2013robust}.
The RMSE of more outdoor experiments is shown in Table.~\ref{tab:outdoor_experiment}.
Our system achieved the best performance in all sequence.

We also perform a larger outdoor experiment and compared the results with Google map. 
Estimated trajecoties are shown in Fig.~\ref{fig:pathonmap}.
The estimated trajecotry of VIO drifted along with distance. 
Thanks to the global sensors, the trajectory of proposed system is almost drift-free, which matches the road map very well.

\section{Conclusion}
\label{sec:conclusion}
In this paper, we propose a optimization-based framework to fuse local estimations with global sensors. 
Local estimations came from previous VO/VIO work.
Global sensors are treated as general factors in pose graph optimization.
This system achieves locally accurate and globally drift-free pose estimation. 
We demonstrate the impressive performance of our system on public datasets and with real-world experiments.

\clearpage
\bibliography{paper.bib}

\end{document}